\documentclass[10pt,twocolumn,letterpaper]{article}

\usepackage{iccv}
\usepackage{times}
\usepackage{epsfig}
\usepackage{graphicx}
\usepackage{amsmath}
\usepackage{amssymb}
\usepackage{color}
\usepackage{multirow}
\usepackage{soul}
\usepackage{floatrow}
\usepackage{caption}
\usepackage{ctable}
\usepackage[symbol]{footmisc}
\usepackage{enumitem}
\usepackage{subfigure}

\usepackage[pagebackref=true,breaklinks=true,letterpaper=true,colorlinks,bookmarks=false]{hyperref}

\iccvfinalcopy 

\newcommand{\ours}{HACS\xspace}

\def\iccvPaperID{964} 

\ificcvfinal\fi
\begin{document}

\title{
    HACS: Human Action Clips and Segments Dataset \\ 
    for Recognition and Temporal Localization
}

\author{
    Hang Zhao$^\dagger$, Antonio Torralba$^\dagger$, Lorenzo Torresani$^\ddagger$, Zhicheng Yan$^\flat$\\
    $^\dagger$Massachusetts Institute of Technology, $^\ddagger$Dartmouth College, $^\flat$University of Illinois at Urbana-Champaign \\
}




\maketitle

\begin{abstract}
    This paper presents a new large-scale dataset for recognition and temporal localization of human actions collected from Web videos. We refer to it as \textbf{\ours} (\textbf{H}uman \textbf{A}ction \textbf{C}lips and \textbf{S}egments). 
    We leverage both consensus and disagreement among visual classifiers to automatically mine candidate short clips from unlabeled videos, which are subsequently validated by human annotators. 
    The resulting dataset is dubbed \textbf{\ours Clips}.
    Through a separate process we also collect annotations defining action segment boundaries.
    This resulting dataset is called \textbf{\ours Segments}. Overall, \textit{\ours Clips} consists of 1.5M annotated clips sampled from 504K untrimmed videos, and \textit{\ours Segments} contains 139K action segments densely annotated in 50K untrimmed videos spanning 200 action categories. \ours Clips contains more labeled examples than any existing video benchmark. This renders our dataset both a large-scale action recognition benchmark and an excellent source for spatiotemporal feature learning. In our transfer learning experiments on three target datasets, \ours Clips outperforms Kinetics-600, Moments-In-Time and Sports1M as a pretraining source.
    On \ours Segments, we evaluate state-of-the-art methods of action proposal generation and action localization, and highlight the new challenges posed by our dense temporal annotations.
\end{abstract}


\section{Introduction}


%
%
%


Recent advances in computer vision~\cite{he2017mask,He2016CVPR} have been fueled by the steady growth in the scale of datasets.
For image categorization, in the span of just a few years we transitioned from Caltech101~\cite{fei2006one}, which contained only 9.1K examples, to the ImageNet dataset~\cite{deng2009imagenet}, which includes over 1.2M examples.
In object detection, we have seen a similar trend in scaling-up dataset sizes.
Pascal VOC~\cite{Everingham10} was first released with 1.6K examples, while the COCO dataset~\cite{lin2014microsoft} today consists of 200K images and 500K object-instance annotations. Open Images V4~\cite{krasin2016openimages} further scales up the size of image datasets to the next level. 
It currently contains 9M images with image-level label and 1.7M images with 14.6M bounding boxes, and has greatly pushed the advances of research work in those fields~\cite{akiba2018pfdet, gaoapproach}.


\begin{figure}[t]
    \vspace{-1em}
    \includegraphics[width=1.0\linewidth]{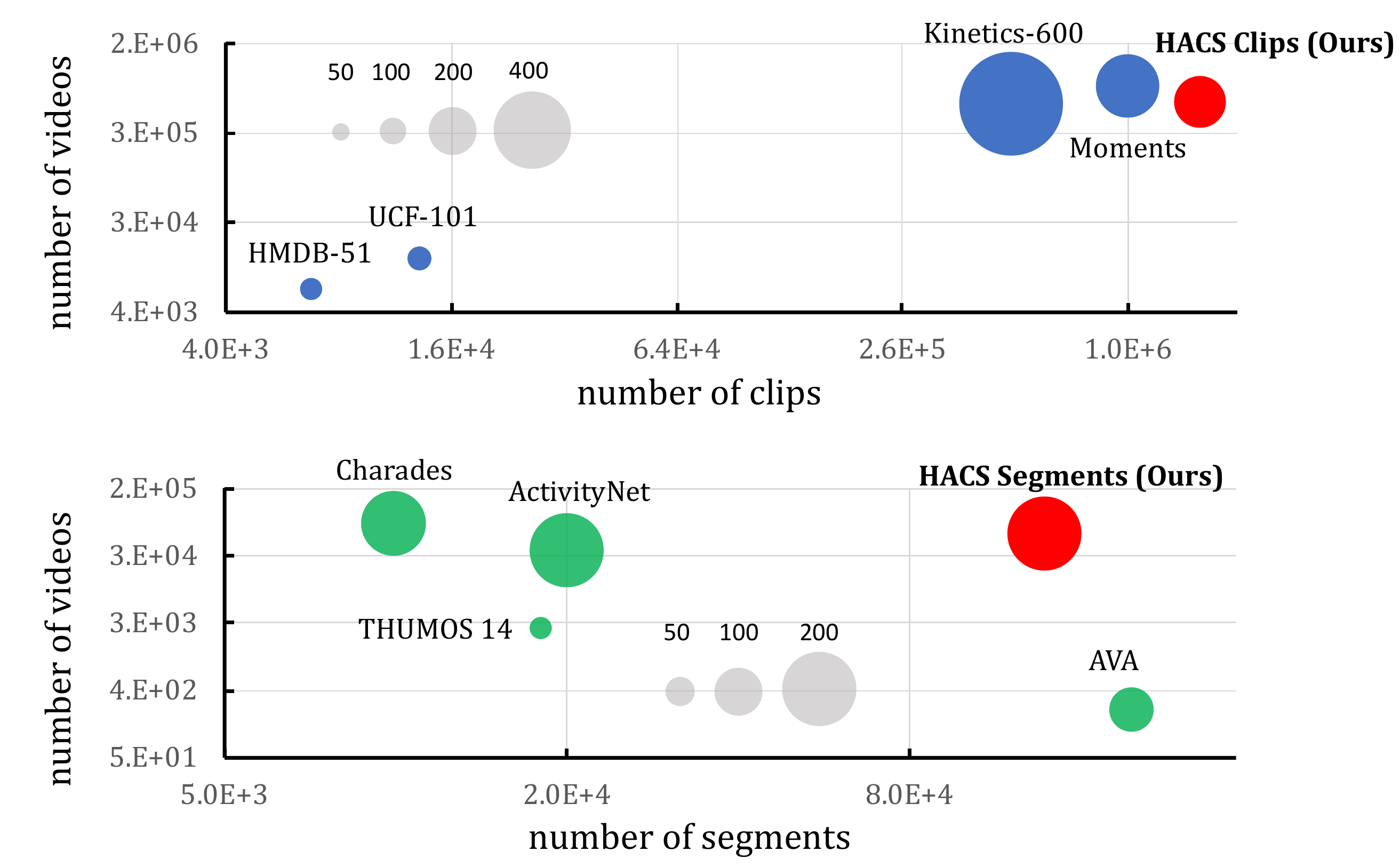}
    \caption{Comparisons of manually labeled action recognition datasets (\textbf{Top}) and  action localization datasets (\textbf{Bottom}), where ours are marked as red. The marker size encodes the number of action classes in logarithmic scale.}
    \label{fig:dataset comparison}
    \vspace{-1em}
\end{figure}

In the video domain, we have witnessed an analogous growth in the scale of action recognition datasets. While video benchmarks created a few years ago consists of only a few thousands examples (7K videos in HMDB51~\cite{kuehne2011hmdb}, 13K in UCF101~\cite{soomro2012ucf101}, 3.7K in Hollywood2~\cite{marszalek2009actions}), more recent action recognition datasets, such as Sports1M~\cite{karpathy2014large}, Kinetics~\cite{kay2017kinetics} and Moments-in-Time~\cite{monfort2019moments}, include two orders of magnitude more videos. 
However, for action localization, we have not seen a comparable growth in dataset sizes. THUMOS~\cite{jiang2014thumos} was created in 2014 and contains 2.7K untrimmed videos with localization annotations over 20 classes. ActivityNet~\cite{caba2015activitynet} only includes 20K videos and 30K annotations. AVA~\cite{pantofaru2017ava} includes 58K clips, and Charade~\cite{sigurdsson2016hollywood} contains 67K temporally localized intervals. We argue that the lack of large-scale action localization datasets is impeding the exploration of more sophisticated models.

Motivated by the needs of large-scale action datasets, we introduce a new video benchmark, named {\em Human Action Clips and Segments} (\textit{\ours})\footnote{Homepage: \url{http://hacs.csail.mit.edu}}. It includes two types of manual annotations. The first type is action labels on \textit{1.5M} clips of 2-second duration sparsely sampled from a half million of videos. We refer to this dataset as \textit{\ours Clips}. It is designed to serve as a benchmark and as a pretraining source for action recognition. In our empirical study we compare different clip sampling methods and we observe that both consensus and disagreement over different visual classifiers can be used as criteria to identify clips especially worthy of annotation. Clips sampled from a large pool of videos according to such criteria capture large variations in action dynamics, context, viewpoint, lighting and other conditions. 
We demonstrate that spatiotemporal features learned on \textit{\ours Clips} generalize well to other 
datasets.

The second type of annotation involves temporal localization labels on \textit{50K} untrimmed videos, where both the temporal boundaries and the action labels of action segments are annotated. We call it \textit{\ours Segments}. Thanks to our stringent guidelines on how to distinguish action and non-action segments, the resulting dataset has 1.8$\times$ more action segments per video and segments of shorter duration compared to ActivityNet. We demonstrate that this poses bigger challenges in action localization, as localizing short segments requires finer temporal resolution and more discriminative feature representations. 
Both types of annotation share the same taxonomy of 200 action classes, which we take from ActivityNet. We compare \ours with other video datasets in Figure~\ref{fig:dataset comparison}. Despite being in its very first version, \ours compares favorably in scale to most prior benchmarks in this area. In summary, we make the following contributions in this paper.
\begin{enumerate}[topsep=1pt, partopsep=0pt, itemsep=2pt, parsep=2pt]
   \item We present a thorough empirical study on clip sampling methods, and use the nontrivial findings to sample a large number of clips for further manual verification. The resulting \textit{\ours Clips} dataset has 2.5$\times$ more clip annotations compared to Kinetics-600.
    \item We benchmark state-of-the-art action recognition models on \textit{\ours Clips}. We show that \textit{\ours Clips} outperforms Kinetics-600, Moments-In-Time and Sports1M as a pretraining dataset for 
   action recognition on other benchmarks.   
   \item We collect action segment boundaries on 50K videos, based on annotation guidelines that reduce the ambiguity in the action definition and localization. The resulting \textit{\ours Segments} has 2.5$\times$ more videos and 4.7$\times$ more action segments compared to ActivityNet. 
   \item On \textit{\ours Segments}, we evaluate state-of-the-art methods of both action proposal generation and action localization, and highlight the new challenges.
\end{enumerate}

\section{Related Work}
\label{sec:related_work}

\noindent \textbf{Action Recognition}. 
In action recognition, the HMDB51~\cite{kuehne2011hmdb} and the UCF101~\cite{soomro2012ucf101} datasets were created to provide benchmarks with higher variety of actions compared to precedent datasets, such as KTH~\cite{schuldt2004recognizing}. These benchmarks have
have enabled hand-design of motion features such as Spatial-Time Interest Point~\cite{laptev2005space}, Spatiotemporal Histogram of Oriented Gradient and Optical Flow~\cite{wang2011action, laptev2008learning} and Fisher Vector feature encoding~\cite{oneata2013action}.
However, these datasets are not large enough to support modern end-to-end training of deep models. The large-scale Sports1M~\cite{KarpathyCVPR14}  and Kinetics datasets~\cite{kay2017kinetics}, which are over 20$\times$ larger than UCF101, were recently introduced to fill this gap. They enable the training of deep models from scratch~\cite{Carreira_2017_CVPR, qiu2017learning, tran2017closer}.
However, these benchmarks cannot be used to train action localization models as they do not contain temporal boundary annotation. Collecting annotations on large-scale video datasets is time-consuming~\cite{sigurdsson2016much}.
Previous work~\cite{ma2017less, li2017attention} have shown that Web action images, which are widely available, can be exploited to train action classifiers, but such images cannot be used to learn motion features. Researchers have also explored synthetic generation of videos (e.g. VGAN~\cite{vondrick2016generating}, PHAV~\cite{de2017procedural} for training action recognition models. Although this eliminates the need for human annotation, models trained on synthetic videos are still inferior to those trained on natural videos with human annotation.


\noindent \textbf{Action Localization}. Action localization in untrimmed videos is crucial to understanding Internet videos.
Recently, several datasets for have been presented. 
THUMOS Challenge 2014~\cite{jiang2014thumos} includes 2.7K trimmed videos on 20 actions. It was subsequently extended into MultiTHUMOS~\cite{yeung2015every} to have 65 action classes.
Other datasets with fine granularity of classes but focused on narrow domains include MPII Cooking~\cite{MPIICooking:CVPR2012, MPIICooking2:IJCV2015}
and EPIC-Kitchens~\cite{Damen2018EPICKITCHENS}.
Unfortunately models trained on such domain-focused datasets may not generalize well to every-day activities. Conversely, the Charades dataset~\cite{sigurdsson2016hollywood} was purposefully designed to include more general, daily activities. 
ActivityNet-v1.3~\cite{caba2015activitynet} includes 20K untrimmed videos and 30K temporal action annotations. More recently, the AVA dataset~\cite{AVA} was introduced to provide person-centric spatiotemporal annotations on atomic actions. 
These datasets have substantially advanced the progress of research on action proposal generation~\cite{gao2018ctap, gao2017turn, buch2017sst, lin2018bsn} and action localization~\cite{xu2017r, shou2017cdc, SSN2017ICCV, lin2017single, chao2018rethinking, bai2018contextual, sstad_buch_bmvc17}. 

\section{Dataset Collection}
\label{sec:data_collect}

\subsection{\ours Dataset at a Glance}
\label{sec:SDAC_overview}
\ours uses a taxonomy of 200 action classes, which is identical to that of the ActivityNet-v1.3 dataset. It has 504K videos retrieved from YouTube.  Each one is strictly shorter than 4 minutes, and the average length is $2.6$ minutes. A total of $1.5$M clips of 2-second duration are sparsely sampled by methods based on both uniform randomness and consensus/disagreement of image classifiers. $0.6$M and $0.9$M clips are annotated as positive and negative samples, respectively. We split the collection into training, validation and testing sets of size 1.4M, 50K and 50K clips, which are sampled from 492K, 6K and 6K videos, respectively. We refer to this benchmark as \textit{\ours Clips}. Furthermore, on a subset of 50K videos (38K for training, 6k for validation and 6K for testing) we collect manual boundaries defining the start, the end and the action label of every action segment in the video. All videos contain at least one action segment. We refer to this collection as \textit{\ours Segments}.


\subsection{Video Retrieval and De-duplication}
We use 200 action labels to query the YouTube video search engine, and retrieve 890K potentially-relevant videos. The number of videos per class ranges from 1100 to 6600. We then perform two types of de-duplications. First, duplicate videos within \textit{\ours} are removed. Second, to support fair assessment on other benchmarks, we remove videos that overlap with samples in the validation or test sets of other datasets, including Kinetics, ActivityNet, UCF-101 and HMDB-51. More details of video de-duplication are included in the supplement.

\subsection{Sparse Clip Sampling}
\label{sec: clip sampling}
Manually annotating the start and the end of action segments in untrimmed videos is time-consuming. 
If the objective is to create a dataset for action recognition, it is more efficient to sparsely sample clips of short duration from a large number of videos, and ask annotators to quickly verify whether the presumed action is truly happening in the clip. This procedure can be used to gather a large-scale action clip dataset that can not only serve as an action recognition benchmark alone, but can also be leveraged for transfer learning, \eg, by enabling the training of general deep models that can then be  transferred for finetuning on smaller-scale datasets or employed in other downstream tasks.

One challenge in sampling clips is that the frequency of positive examples is arguably much smaller than that of negative examples. Thus, uniform random clip sampling would inevitably yield a large number of negative examples which are far less useful than positive examples for video modeling. 
On the other hand, using machine learning classifiers to guide the clip sampling can introduce dataset biases. For example, the collection of Kinetics~\cite{kay2017kinetics} clips leveraged an image classifier trained on images automatically labeled by user feedback from Google Image Search. This classifier was used to sample clips with top action scores. The bias induced by such image classifier is certainly present in the data, yet it is difficult to assess.

\begin{figure}[t]
\centering
\centerline{\includegraphics[width=1.0\linewidth]{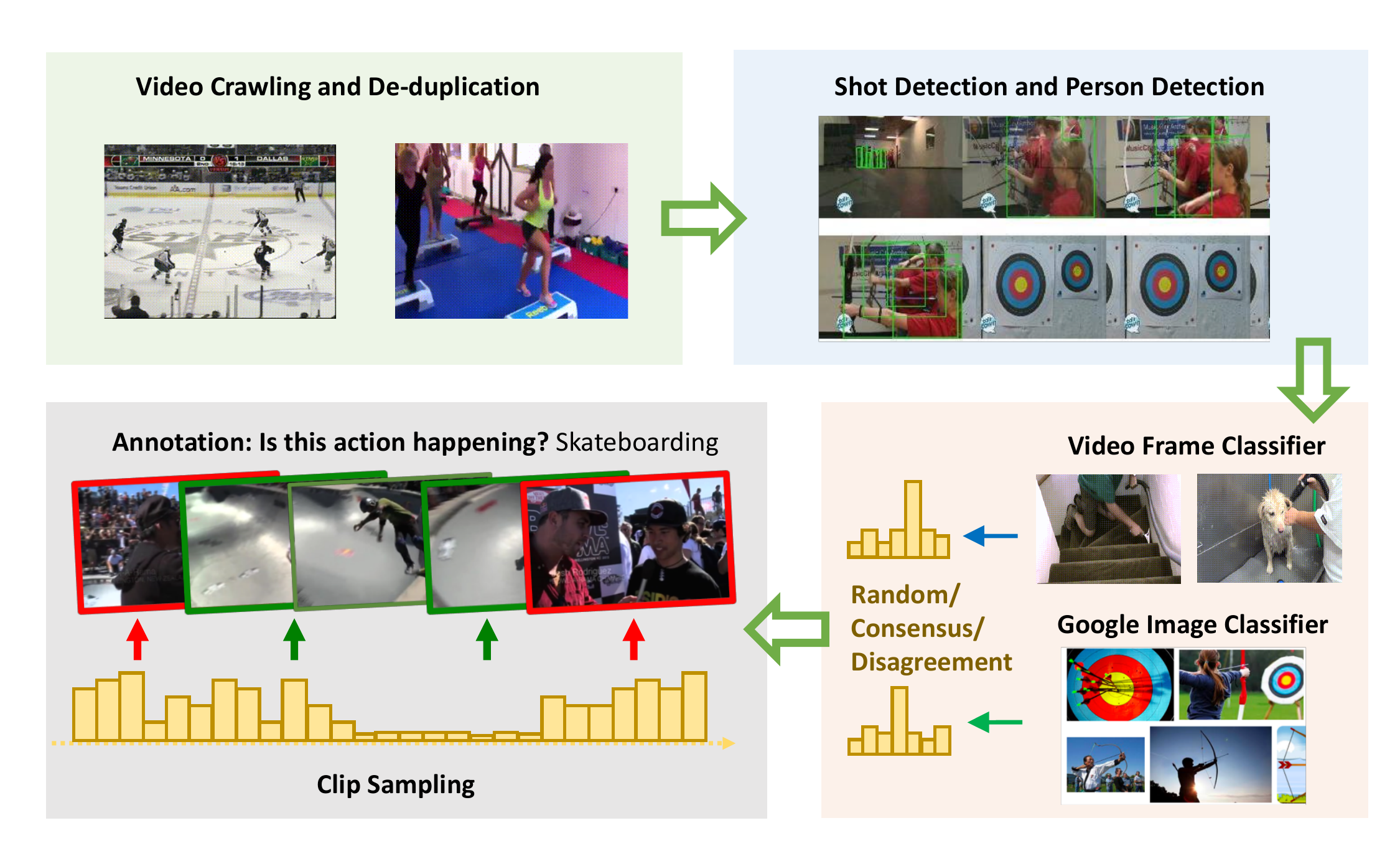}}
   \caption{Our pipeline of sparse clip sampling and labeling.}
   \label{fig:sparse clip pipeline}
\end{figure}

In this section we are interested in the following two questions. First, \textit{how can we assess the quality of clips sampled by different methods}? Second, \textit{which clip-sampling method gives rise to the best training dataset}? To answer these questions, we present a thorough empirical study of clip-sampling strategies. An overview of the clip sampling pipeline used in our study is shown in Figure~\ref{fig:sparse clip pipeline}.

\subsubsection{Preprocessing: Removing Non-Person Clips}
As a preprocessing step, we exclude clips that do not contain people since our aim is to create a dataset of human actions. To accomplish this, we first run a shot detection based on color histogram distance between video frames to segment the video into shots. After that we run a Faster R-CNN~\cite{ren2015faster} person detector on two frames uniformly spaced in each shot, and remove shots with low average person detection scores. 

\subsubsection{A Study on Clip Sampling Methods}
\label{sec: study clip sampling}
In this study, we compare three sampling methods: random sampling and two image classifier-based sampling methods. Prior work ~\cite{ma2017less, li2017attention, guo2014survey} on exploiting still images for action recognition has shown that still-image classifiers can predict actions in video reasonably well, despite their inability to model motion. Action context, such as objects typically involved in the action, prototypical scenes where actions occur, and other visual patterns that frequently co-occur with the action, can be captured by the image classifier for recognizing actions. To support our study, we first train two distinct image classifiers using training data from two different domains:
\begin{itemize}[noitemsep,topsep=0pt,leftmargin=*]
\item \textbf{YouTube Frame Model}. The first model is trained on frames extracted from the top-500 videos retrieved by YouTube for each action class. 
Only video frames with person detected are used as positive samples for training. This gives a total of over 600K frames. 
As background (negative) samples we randomly choose frames with low person-score.
\item \textbf{Google Image Model}. The second model is trained on images retrieved from the Google Image Search engine using the class labels as queries. We collect a total of 304K images after thresholding on person detection score. We use random samples from ImageNet as the examples of the background class.
The image distribution is different from that of video frames, in terms of scene composition, background, and viewpoint.
\end{itemize}

\noindent For both classifiers we use a ResNet-50 trained with cross-entropy loss over 201 classes (200 action classes and 1 background class). The classifiers are applied to the central frame in each shot to get a probabilistic action prediction.






\noindent Next, we consider three different clip sampling methods:
\begin{enumerate}[noitemsep,topsep=0pt,leftmargin=*]
\item \textbf{Random}. We randomly sample frames from each video. 

\item \textbf{Maximum Entropy (ME)}. Within each video, we define the unnormalized sampling probability for the central frame of each shot as the average entropy of probabilistic predictions from the two image classifiers. We then apply L1-normalization to obtain a proper sampling distribution over the video. This method prefers to sample frames where the two classifiers disagree the most. 

\item \textbf{Maximum Consensus (MC)}. Different from ME, the MC method defines the unnormalized sampling probability as the average action score from the two image classifiers for the action label that is used to retrieve the video. L1-normalization is also used. This method biases the sampling towards clips that receive a high score from both classifiers for the action label of interest.
\end{enumerate}
Using these 3 sampling strategies, we collect 3 different sets of clips from a subset of training videos, which are denoted as \textit{Train-mini-Random}, \textit{Train-mini-ME} and \textit{Train-mini-MC}, respectively. For each strategy, we randomly select 400 training videos per class, and sample 3 frames per video. Clips of 2-second duration centered around these sampled frames are sent to human annotators for manual verification, and each clip is marked as either positive or negative w.r.t the label of interest. Most action classes in our taxonomy are sufficiently distinct when observed in 2-second clips and annotating 2-second clips is also efficient.

\begin{table}[t!]
    \begin{tabular}{c|ccc}
      Clip Type & ME & Random & MC \\ 
       \specialrule{.15em}{.1em}{.1em}      
      Positive clips  & 71.3K & 82.2K & \textbf{100.3K}\\ 
      Negative clips & \textbf{168.7K} & 157.8K & 139.7K \\
    \specialrule{.1em}{.05em}{.05em}
    \end{tabular}

  \caption{Comparing the frequency of positive and negative clips in three \textit{Train-mini} sets sampled by different methods. 
  }
  \label{tab:clip stat} 
\vspace{-2pt}  
\end{table}

\begin{figure}[ht!]
\centering
\centerline{\includegraphics[width=1.\linewidth]{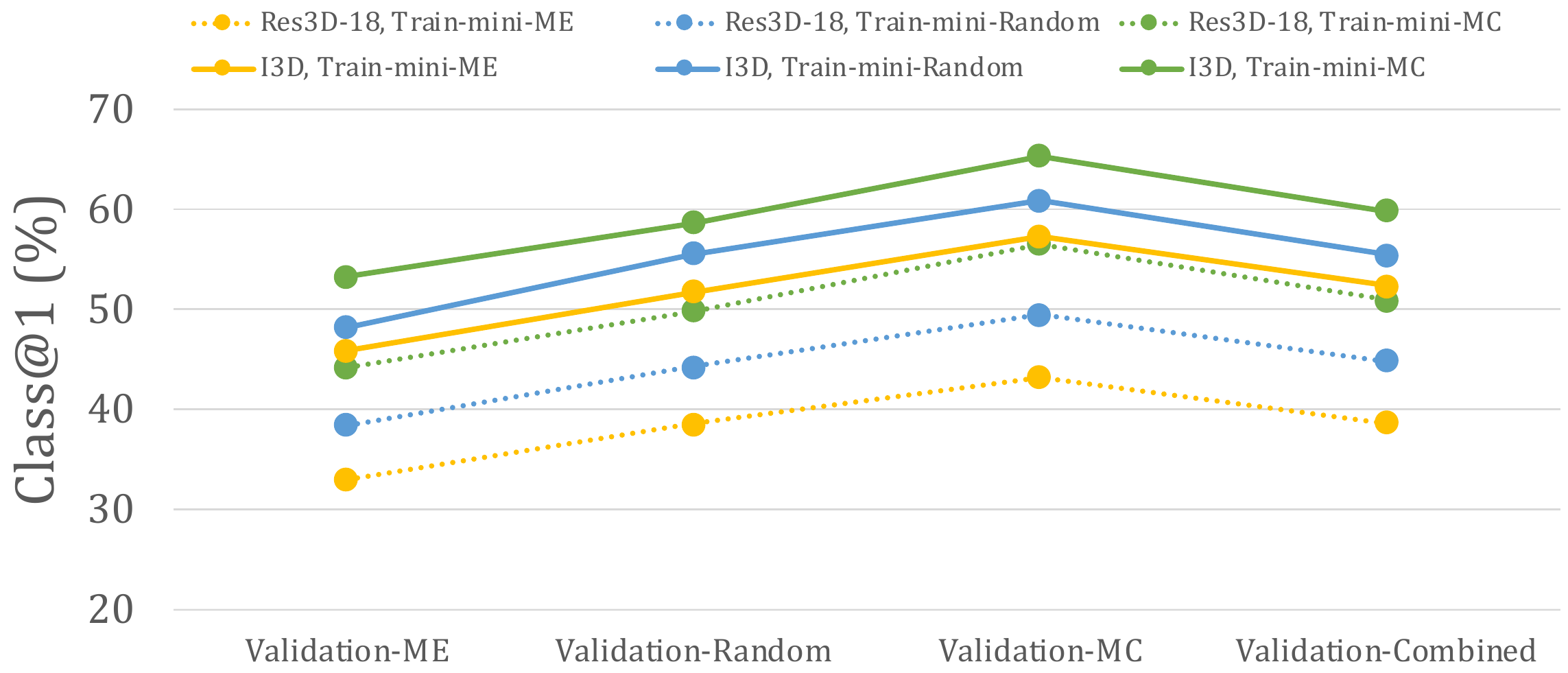}}
  \caption{Evaluating Res3D-18 and I3D models trained on 3 different \textit{Train-mini} sets on 4 different validation sets.}
  \label{fig:sampling study}
\end{figure}

\noindent \textbf{Statistics of sampled clips}. As shown in Table~\ref{tab:clip stat}, MC method samples the highest number of true positive clips since clips with high scores based on the consensus of image classifiers are more likely to be true positive. However, these are also likely to be easy positive examples as they can be recognized by image classifiers. On the other side, ME yields the smallest number of true positives since it samples clips with conflicting predictions from image classifiers. This implies more uncertainty about the action class.

\begin{figure*}[th]
\vspace{-5em}

\begin{tabular}{c}
    \includegraphics[width=0.8\linewidth]{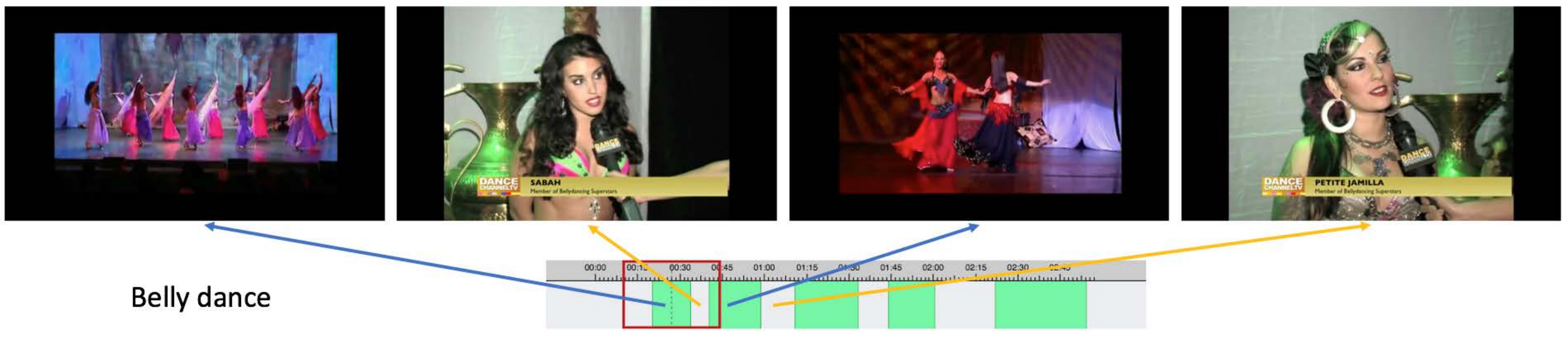} \\ \\
    \includegraphics[width=0.8\linewidth]{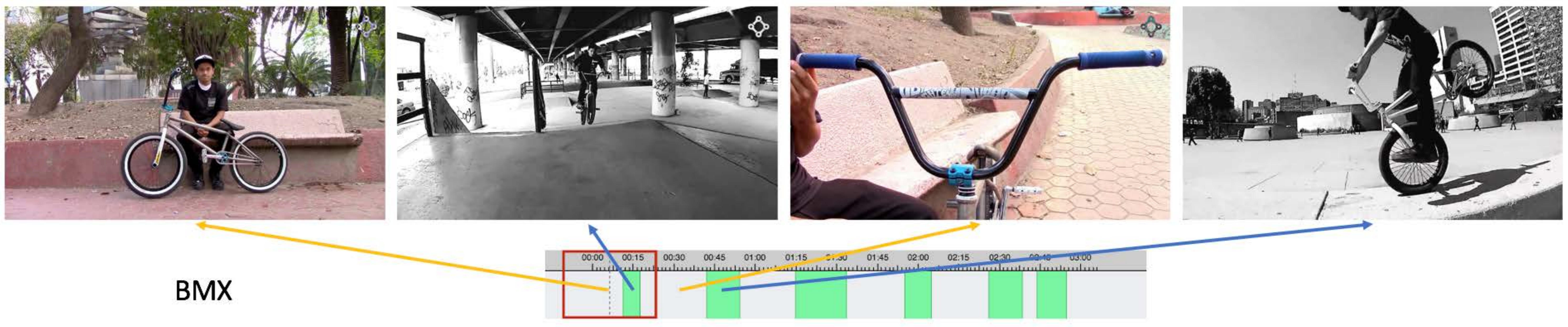} \\ \\
    \includegraphics[width=0.8\linewidth]{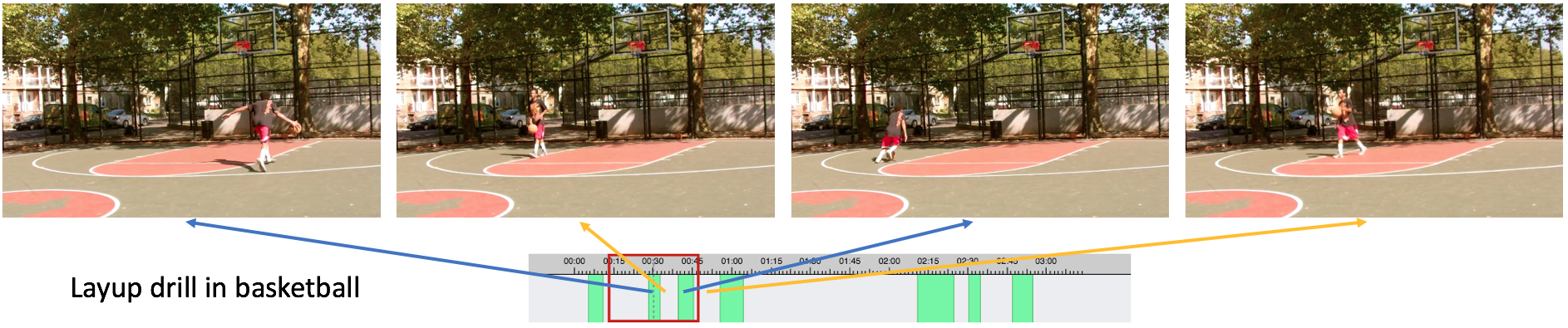}
\end{tabular}
\caption{Examples of dense segment annotations. Action definition is clarified in the guideline to reduce the ambiguity of action boundaries.}
\label{fig:dense annotation example}

\vspace{-0.5em}
\end{figure*}

\noindent \textbf{Evaluating clip sampling methods.} We perform an empirical evaluation to address the two questions we asked in Section~\ref{sec: clip sampling}.  Two models are used. A Res3D model~\cite{tran2017closer} with 18 residual units (\ie \textit{Res3D-18}) and a I3D model~\cite{Carreira_2017_CVPR}. Both take sequences of 16 frames as input. At training time, a random sequence of 16 frames within the clip is used. At evaluation time, 4 evenly spaced sequences of 16 frames are used and their predictions are averaged to obtain the final prediction.  
We train 3 separate instances of each model on the 3 different \textit{Train-mini} sets. Since positive and negative clips are imbalanced, we adopt weighted sampling during training where the weight of each example is inversely proportional to the square root of the size of its class. 

We also apply each sampling method to validation videos, and obtain 3 different sets of clips, namely \textit{Validation-Random}, \textit{Validation-ME} and \textit{Validation-MC}, respectively, They are also manually verified by humans. We also combine all of 3 validation sets into a single one, namely \textit{Validation-Combined}. We evaluate each trained model instance on all of 4 validation sets. Since the validation sets are also class-imbalanced, we report mean class accuracy (Class@1), which is obtained by averaging per-class accuracy over the 201 classes. 

The results are shown in Figure~\ref{fig:sampling study}. Models trained on the \textit{Train-mini-MC} set consistently outperforms models trained on \textit{Train-mini-Random} and \textit{Train-mini-ME} sets on all validation sets. This suggests that for constructing a large-scale training set of clips under a constant human annotation budget, MC is the best method among those considered here because models trained on \textit{Train-mini-MC} generalize best to all types of validation sets. On the other side, the \textit{Validation-MC} set is easier than the others (models achieve higher accuracy), while \textit{Validation-ME} is the most difficult for all models. This indicates that to construct a less biased validation/testing set, we should not rely on a single sampling method. Therefore, we propose to combine clips sampled by all of 3 methods in the final validation/testing set.

\subsection{Sparse Clip Annotation}
\label{sec:clip_annotation}
We set up annotation tasks to label the sampled clips. 

\noindent \textbf{Annotation Guideline}. 
Different people may have different understandings of what constitutes a given action. 
To reduce the ambiguity, we prepare a detailed annotation guideline, which includes both clear action definitions as well as positive/negative examples with clarifications separately for each action. See more detailed guideline in the supplement.


\noindent \textbf{Annotation Tool}. Our annotation tool supports display of up to 200 clips in a single page. We present clips sampled from the same video together. This not only reduces annotation inconsistency but also makes the annotation faster.  

\noindent \textbf{Quality Control}.
We make two efforts to improve the annotation quality. First of all, each clip is labeled by three annotators, and only those clips with consensus from at least two annotators are included in the final dataset. Secondly, we ensure clips from the same class are labeled by the same group of annotators. This removes the inter-annotator noise.

\begin{figure}[ht!]
	\centering
	\includegraphics[width=0.9\linewidth]{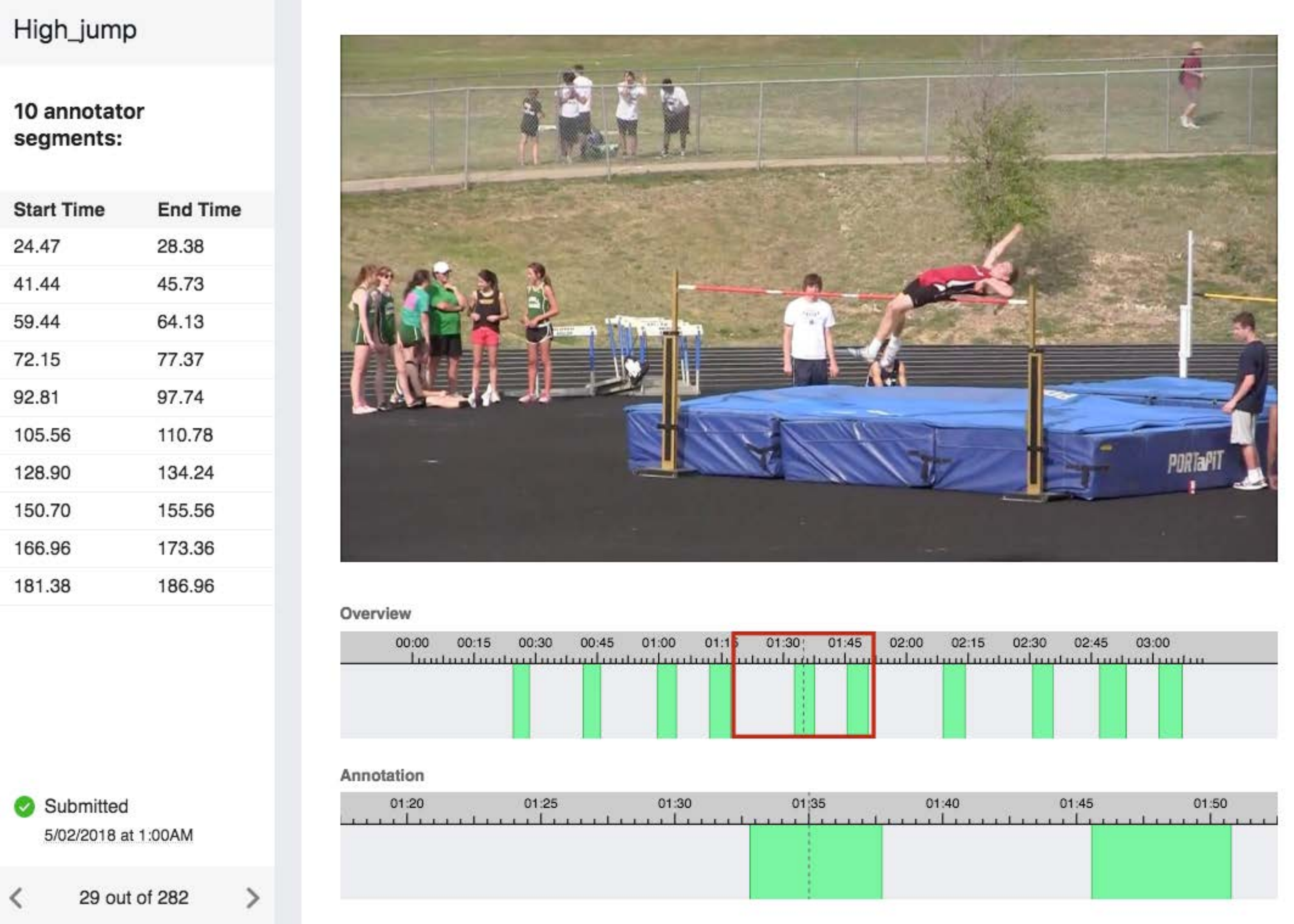}
	\caption{Action segment annotation tool. A timeline overview is shown below the video player, and a zoom-in view of current time window is shown in the bottom for accurate temporal annotation.}
	\label{fig:dense_segment_annotation}
\vspace{-0.5em}
\end{figure}

\begin{figure*}[ht!]
\vspace{-5em}
    \begin{subfigure}
        \centering
        \includegraphics[width=1.0\linewidth]{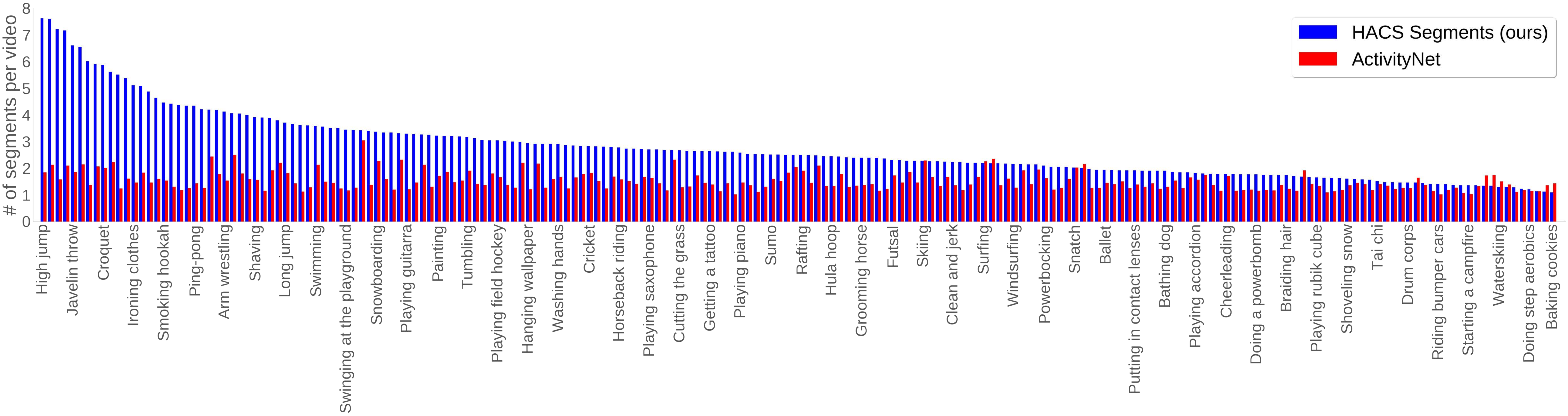}
    \end{subfigure}

\vspace{-2.1em}

    \begin{subfigure}
        \centering
        \includegraphics[width=1.0\linewidth]{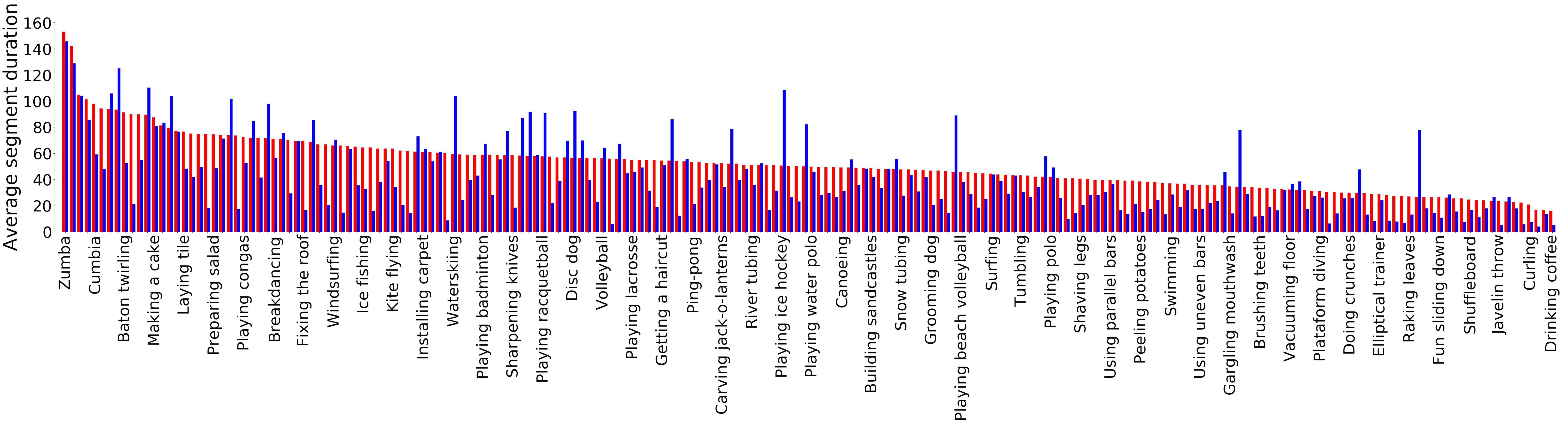}
    \end{subfigure}

	\caption{Comparing \ours Segments and ActivityNet. \textbf{Top}: comparing average number of action segments per video. On average, \ours Segments has 1.8$\times$ action segments per video (2.8 Vs. 1.5 segments). \textbf{Bottom}: comparing average segment duration. \ours segments are significantly shorter than those in ActivityNet (40.6 Vs. 51.4 seconds). \vspace{-0.5em}}
	\label{fig:compare sdac and actnet}
\end{figure*}

\subsection{Dense Segment Annotation}

\ours Clips alone are not sufficient for training and evaluating action localization methods as they lack temporal boundaries. Therefore, we ask annotators to densely label the start, the end and the action class of all action segments in a subset of 50K videos. A screenshot of our dense segment annotation tool is shown in Figure~\ref{fig:dense_segment_annotation}. We prepare clear annotation guidelines on distinguishing foreground action segments, where the action is being performed, and background segments, where both the person and the context (\eg objects, scene) may appear but the action is not present. More importantly, we identify common patterns of the start and end of each action class. This helps annotators to better annotate the action segment boundaries. Examples of dense segment annotations are shown in Figure~\ref{fig:dense annotation example}. For instance, for action \textit{Belly Dance}, we consider the part of the video where dancers are being interviewed as background. For action \textit{BMX}, we suggest to mark as background the part of the video where the person is explaining how to ride BMX bikes  even though the rider and BMX bikes are visible. For action \textit{Layup drill in backetball}, we clarify that the part of the video where the player stands still or has finished the shooting should be marked as background. 

The efficacy of our guidelines can be measured in numbers: compared to ActivityNet, \ours has on average 1.8$\times$ more action segments per video, and the average segment duration is about $20\%$ shorter, as shown in Figure~\ref{fig:compare sdac and actnet}.  This poses new challenges to action localization methods, which have to localize more segments of shorter duration.

\begin{table}[t!]
\begin{tabular}{c|c c c}
    Input & RGB & Flow & RGB+Flow \\ 
	\specialrule{.15em}{.1em}{.1em}
	Class@1 & 80.3 & 72.2 & 83.5\\ 
	\specialrule{.1em}{.05em}{.05em}
\end{tabular}
  \caption{Evaluating I3D models~\cite{Carreira_2017_CVPR} on the validation set of \ours Clips.\vspace{-1.5em}}%
  \label{tab: hacs_clips_eval}
\end{table}

\subsection{Distinguishing Properties of \ours}
Unlike other recognition datasets where only a single positive example is collected per video, \ours Clips includes also negative examples (each video contains 3 clips, with the negative to positive ratio being roughly 1 to 2). 
This could be used to model the discrepancy between action and non-action content. Moreover, videos in \ours Segments have both sparse clip annotation, which is a weak form of supervision for localization~\cite{Wang_2017_CVPR, shou2018autoloc, nguyen2018weakly}, and dense segment annotation, which can serve as the ground-truth of localization. Such hybrid annotation can be used for the task of weakly supervised action localization~\cite{Wang_2017_CVPR, shou2018autoloc}, reminiscent of point supervision~\cite{bearman2016s} and scribble supervision~\cite{lin2016scribblesup} in image semantic segmentation.



\section{Action Recognition on \ours Clips}
\label{sec:exp_SDAC}

\subsection{Action Clip Classification}
In this section, we train I3D~\cite{Carreira_2017_CVPR} on the full \ours Clips training set, and evaluate it on the validation set. 
We experiment with both RGB frames and optical flow as input. For efficiency, Farneback's algorithm~\cite{farneback2003two} is adopted to compute optical flow. We also report results of combining RGB and optical flow by late fusion, where the final prediction score is a weighted sum of the prediction scores obtained from RGB and optical flow. We empirically set fusion weights for RGB and optical flow to 0.6 and 0.4, respectively. The results are shown in Table \ref{tab: hacs_clips_eval}. 
We also show the class-specific accuracy, as well as the distribution of positive and negative clips per class in the supplement. 


\subsection{Results of Transfer Learning}
\label{sec:clip transfer learning}

Models trained on \ours Clips can be finetuned on other recognition datasets.
By comparing finetuned models with models trained from scratch, we can assess the generalization performance of spatial-temporal features learned on \ours Clips. We evaluate the transfer learning on 3 action recognition benchmarks. On all benchmarks we observe substantial gains by pre-training on \ours Clips. 

\begin{table}[t]
	\begin{center}
        \resizebox{\columnwidth}{!}{	
    		\begin{tabular}{c|c|ccc}
    			Input & Pretraining & UCF101 & HMDB51 & Kinetics400  \\ 
    			\specialrule{.15em}{.1em}{.1em}
    			\multirow{4}{*}{RGB}& None  & 75.0 & 39.4 & 69.9 \\ 
    			& Sports1M & 92.8 & 68.3 & 71.0  \\
    			& Moments & 92.4 & 69.6 & 71.6  \\
    			& Kinetics-600 & 94.9 & 73.4 & 72.9  \\
    			& \ours Clips & \textbf{95.1} & \textbf{73.6} & \textbf{73.4} \\ \hline
    			\multirow{4}{*}{Flow}& None & 85.2 & 56.1 & 62.9 \\ 
    			& Sports1M & 92.7 & 71.1 & 63.4 \\
    			& Moments & 94.6 & 75.3 & 63.9 \\
    			& Kinetics-600 & \textbf{96.0} & 76.2 & 66.7 \\
    			& \ours Clips & 95.7 & \textbf{76.5} & \textbf{67.2} \\
    			\specialrule{.1em}{.05em}{.05em}
    		\end{tabular}
        }
	\end{center}
	\vspace{-1em}
	\caption{Comparisons of \ours Clips with other datasets for pre-training I3D models. Results of UCF-101 and HMDB-51 are computed on split 1. \textit{Moments} denotes Moments-In-Time dataset.
	}
	\label{tab: pretrain_dataset_comp}
\end{table}


\noindent \textbf{Datasets}. We use  a total of 6 additional datasets for our assessment. UCF-101, HMDB-51 and Kinetics-400 are used as target benchmarks. Sports1M, Moments-in-Time and Kinetics-600~\cite{kinetics600}, which is an extended version of the original Kinetics-400 dataset, are used as comparative pre-training datasets. 
For Kinetics-400, we report the accuracy on the validation set. 
For evaluation metric, we use Video@1 which is obtained by evenly sampling 10 clips in the video, and averaging the predictions.

\noindent \textbf{Results}. We train I3D models~\cite{Carreira_2017_CVPR} without any use of 2D-to-3D inflation. When I3D models are pre-trained, we further fine-tune them on target benchmarks.
As shown in Table \ref{tab: pretrain_dataset_comp}, by pretraining on \ours Clips, the metrics are substantially improved on all 3 benchmarks. 
On all target datasets, \ours Clips shows better generalization performance compared to Sports1M, Moments-in-Time and Kinetics-600, where Kinetics-600 is the strongest competitor in this set. 
Sports1M annotations are noisy as they are generated by a tag prediction algorithm. Also, the average length of Sports1M videos is over 5 minutes and the tagged action may only be present for a short period of time. This introduces substantial temporal noise in learning spatial-temporal feature representation. 
Compared to Moments-in-Time, \ours Clips has a more fine-grained taxonomy for human actions, which helps generalization to other datasets.
Compared to Kinetics-600, \ours Clips contains over 3$\times$ more annotations in the training set,
which also contributes to the superior transfer learning performance.

\noindent \textbf{Comparisons with other methods}. In Table~\ref{tab: state_of_art_comp} we compare with the state-of-the-art. Both I3D~\cite{Carreira_2017_CVPR} and R(2+1)D~\cite{tran2017closer} model architectures are used here. For R(2+1)D, we report results of models with both 34 and 101 residual units after late fusion of RGB and flow scores. We compute video classifications by averaging predictions over 20 evenly-sampled clips in each video. By using the off-the-shelf I3D and R(2+1)D models, and leveraging a large-scale clip dataset, our approach outperforms other methods~\cite{varol2017long,feichtenhofer2017spatiotemporal,TSN2016ECCV,Carreira_2017_CVPR, tran2017closer} on all 3 benchmarks. 

\noindent \textbf{Transfer learning on action localization}.
\ours Clips can also be used to pretrain action localization models. 
Compared with training from scratch, pretraining CDC models~\cite{shou2017cdc} on \textit{\ours Clips} improves the average mAP by $8.6\%$ on THUMOS 14 and by $2.5\%$ on ActivityNet, respectively. See more detailed results in the supplement.


\begin{table}[t]
	\begin{center}
    \resizebox{\columnwidth}{!}{	
    		\begin{tabular}{c|c | c c c}
    			Pretrain Data & Method & UCF101 & HMDB51 & Kinetics-400  \\ 
    			\specialrule{.15em}{.1em}{.1em}
    			\multirow{3}{*}{ImageNet} & LTC-CNN~\cite{varol2017long} & 92.7 & 67.2 & N/A  \\
    			& ST-Multiplier Net~\cite{feichtenhofer2017spatiotemporal} & 94.2 & 68.9 &  N/A \\
    			& TSN~\cite{TSN2016ECCV} & 94.2 & 69.4 & N/A \\ \hline
    			Sports1M & T-S R(2+1)D-34~\cite{tran2017closer} & 97.3  & 78.7  &  75.4\\ \hline 
    		    Kinetics-400 & T-S I3D~\cite{Carreira_2017_CVPR} & 98.0 & 80.7 & 75.7 \\ \hline
                \multirow{3}{*}{\ours Clips} & T-S I3D & \textbf{98.2} & \textbf{81.3} & 76.4 \\
    			& T-S R(2+1)D-34 & 98.0 & 79.8 & 76.1 \\   
    			& T-S R(2+1)D-101 & N/A & N/A & \textbf{77.0} \\
    			\specialrule{.1em}{.05em}{.05em}
    		\end{tabular}
    }
\end{center}

	\caption{Comparing I3D and R(2+1)D models pretrained on \textit{\ours Clips} with prior work. For UCF-101 and HMDB-51, average results over 3 splits are reported. Because R(2+1)D-101 model has 2$\times$ more residual units and 1.3$\times$ more parameters compared to R(2+1)D-34, it heavily overfits to the small datasets of UCF-101 and HDMB-51. Thus, we omit these results. We use \textit{T-S} to denote Two-Stream.}
	\label{tab: state_of_art_comp}
\end{table}


\section{Action Localization on \ours Segments}
\label{sec:exp_SDAC_segment}

We evaluate two action proposal generation methods and one action localization approach on \ours Segments.

\subsection{Results of Action Proposal Generation}
Two action proposal generation methods are evaluated: Boundary Sensitive Network (BSN)~\cite{lin2018bsn} and Temporal Actionness Grouping (TAG)~\cite{SSN2017ICCV}. We choose them because they achieve SoTA results on THUMOS 14 and ActivityNet benchmarks, and open-source implementations of these methods are available. We mostly follow the original training settings, and only highlight the differences below.


\noindent \textbf{BSN Experiments}. In the original work, snippet-level features are 400D, arising from a concatenation of two 200D 
probability vectors extracted from two TSN~\cite{TSN2016ECCV} models trained on 200 action classes of ActivityNet using RGB input and optical flow input, respectively. Analogously, here we train two TSN models (respectively taking RGB and flow as inputs) on \ours Clips with 200 action classes and 1 background class. The two 201D 
probability vectors from the trained models are concatenated to form 402D snippet-level features.

\noindent \textbf{TAG Experiments}. In the original work, two binary classifiers (based on TSN~\cite{TSN2016ECCV}) are trained on ActivityNet using RGB input and optical flow input, respectively. We use annotation in \ours Segments to train such binary classifiers.

\begin{figure}[t!]
	\centering
	\includegraphics[width=1.0\linewidth]{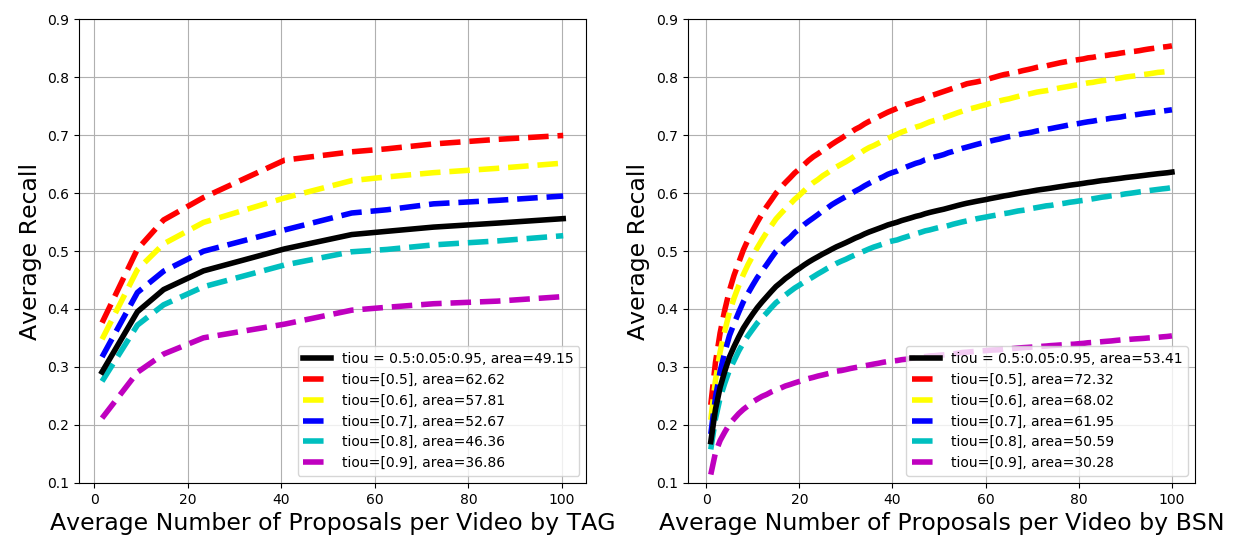}
	\caption{Action proposal generation results of TAG (\textbf{Left}) and BSN (\textbf{Right}) methods on \ours Segments.}
	\label{fig: recall_vs_proposal_number}
\vspace{-0.5em}
\end{figure}

We follow the original evaluation protocols, and report two metrics: 1) Average Recall (AR) vs Average Number (AN) of proposals per video and 2) area under AR-AN curve (AUC). Both are averaged over temporal Intersection over Union (tIoU) thresholds from 0.5 to 0.95 at increments of 0.05. Results are shown in Table~\ref{tab: BSN comparison between SDAC and actnet} (Row 4 \& 5) and Figure~\ref{fig: recall_vs_proposal_number}. Compared to TAG, BSN achieves both better AR@100 and better AUC score. However, TAG achieves higher AUC at high tIoU threshold 0.9 in Figure~\ref{fig: recall_vs_proposal_number}, indicating it is able to better localize action segment boundaries.

\begin{table}[t]
\begin{center}		
	\begin{tabular}{c|c|cc}
		Method & Train/Test Dataset & AR@100 & AUC  \\ 
	    \specialrule{.15em}{.1em}{.1em}
		\multirow{3}{*}{BSN} & ActivityNet & 74.16 & 66.17  \\ 
		& \ours Segments Mini & 61.85 & 51.59  \\  
		& \ours Segments & 63.62 & 53.41  \\ \hline
		TAG & \ours Segments & 55.88 & 49.15  \\ 
		\specialrule{.1em}{.05em}{.05em}
	\end{tabular}		
\end{center}
	\vspace{-0.5em}
	\caption{Action proposal generation results on ActivityNet and \ours Segments. BSN results on ActivityNet are from the original work~\cite{lin2018bsn}. Other results are obtained by running open-source implementations on \ours Segments. }
\label{tab: BSN comparison between SDAC and actnet}
\end{table}

\noindent \textbf{Comparing \ours Segments with ActivityNet}. We use BSN to compare the difficulty of action localization on \textit{\ours Segments} vs ActivityNet. While \textit{\ours Segments} and ActivityNet have validation sets of similar size (6K \textit{vs }5K videos), the training set of \textit{\ours Segments} is 3.8$\times$ larger than that of ActivityNet (38K \textit{vs }10K videos). To have a more fair comparison, we create \textit{\ours Segments Mini}, which contains 10K training videos (50 videos per class) and the original \textit{\ours Segments} validation set. We train and test each model on the training and validation splits of the same dataset (e.g., a model trained on \textit{\ours Segments Mini} is tested only on the validation set of \textit{\ours Segments}, not that of ActivityNet). 

As shown in Table~\ref{tab: BSN comparison between SDAC and actnet} (Row 2 \& 3), compared to ActivityNet, BSN achieves much lower AR@100 and AUC score on \ours Segments Mini. This suggests \ours Segments Mini is a more challenging localization benchmark as it has more segments to localize in each video, and those segments have shorter duration. 
Note we do not experiment with models trained on one dataset and tested on a different one (say, trained on \ours Segments Mini and tested on ActivityNet) as  the definitions where actions start, last and end may vary across datasets. 
Another finding is by training BSN models on the \ours Segments full dataset, AR@100 and AUC are improved by $1.77\%$ and $1.82\%$ in Table~\ref{tab: BSN comparison between SDAC and actnet} (Row 4), which suggests that larger training sets lead to better accuracy.

\noindent \textbf{Exploiting Negative Examples in \ours Clips}. In \ours Clips, we annotated 1M negative clips. Due to the proposed clip sampling method, they include many hard negative examples, such as  clips where both person and context are present, but action is not happening. We have conducted an ablation study on how they can help learn more useful features for action proposal generation. Due to space constraints, the results are presented in supplement.



 \subsection{Results of Action Localization}
We train and
test the Structured Segment Network (SSN)~\cite{SSN2017ICCV} on \ours Segments using its open-source implementation.\footnote{BSN~\cite{lin2018bsn} is not benchmarked because its open-source code does not implement proposal classification.}.
Results are reported in Table~\ref{tab: ssn results}. 
Compared to ActivityNet, localization average mAP on \ours Segments Mini is 12.35\% lower. Given that ActivityNet and \ours Segments Mini have similar numbers and durations of untrimmed videos, the challenging nature of \ours comes from precise segment annotations. 
The average mAP gap between \ours Segments Mini and \ours Segments is 3.04\%. This suggests that the reduction of training data hinders the action localization performance, and that our full-scale training set boosts the accuracy by a large margin.

\begin{table}[t!]
\begin{center}
 	\begin{tabular}{c | c c c c}
 	Dataset & 0.50 & 0.75 &  0.95 & Average \\ 
    \specialrule{.15em}{.1em}{.1em}
 	ActivityNet~\cite{SSN2017ICCV} & 43.26 & 28.70 & 5.63 & 28.28 \\ \hline
 	\ours Segments Mini & 24.89  & 16.04 & 4.50 & 15.93 \\
 	\ours Segments & 28.82 & 18.80 & 5.32 & 18.97 \\
    \specialrule{.1em}{.05em}{.05em}
 	\end{tabular}
 \end{center}
    \vspace{-0.5em}
  	\caption{Action localization results of SSN method for tIoU thresholds ranging from 0.5 to 0.95. Metric is mAP ($\%$) . Results on ActivityNet are from the original work. Results on \ours Segments are obtained by late fusion of scores from RGB and Flow models.}
  	\label{tab: ssn results}
\end{table}


\vspace{-0.5em}
\section{Conclusion}
\label{sec:conclusion}

We introduced a new video dataset with both sparse and dense annotations. We have demonstrated the excellent generalization performance of spatial-temporal feature learned on \textit{\ours Clips} due to its large scale. Compared to other localization datasets, \textit{\ours Segments} is not only larger, but it also poses new challenges in action localization through finer-scale temporal annotations. 
We hope the new challenges in action recognition and localization posed by \textit{\ours} will inspire a new generation of methods and architectures for modeling the high complexity of human actions.



{\small
\bibliographystyle{ieee_fullname}
\bibliography{egbib}
}

\end{document}